\documentclass[letterpaper]{article} 
\usepackage{aaai2026}  
\usepackage{times}  
\usepackage{helvet}  
\usepackage{courier}  
\usepackage[hyphens]{url}  
\usepackage{graphicx} 
\usepackage{natbib}  
\usepackage{caption} 
\usepackage{algorithm}
\usepackage{algorithmic}
\usepackage{booktabs}
\usepackage{amsmath}
\usepackage{amssymb}
\usepackage{multirow}
\usepackage [switch]{lineno}
\usepackage{amsfonts}
\usepackage{pdfpages}
\usepackage{newfloat}
\usepackage{listings}
\DeclareCaptionStyle{ruled}{labelfont=normalfont,labelsep=colon,strut=off} 
\floatstyle{ruled}
\newfloat{listing}{tb}{lst}{}
\floatname{listing}{Listing}
\title{DAPointMamba: Domain Adaptive Point Mamba for Point Cloud Completion}
\author{
    Yinghui Li \textsuperscript{\rm 1}\equalcontrib,
    Qianyu Zhou \textsuperscript{\rm 2}\equalcontrib,
    Di Shao \textsuperscript{\rm 3},
    Hao Yang \textsuperscript{\rm 4},
    Ye Zhu\textsuperscript{\rm 1}\thanks{Corresponding Authors.},
    Richard Dazeley\textsuperscript{\rm 1},\\
    Xuequan Lu\textsuperscript{\rm 5}\footnotemark[2]
}
\affiliations{
    \textsuperscript{\rm 1}School of Information Technology, Deakin University, Australia\\
    \textsuperscript{\rm 2}College of Computer Science and Technology, Jilin University, Jilin, China\\
    \textsuperscript{\rm 3}Department of New Networks, PengCheng Laboratory, Shenzhen, China\\
    \textsuperscript{\rm 4}School of Computer Science, Shanghai Jiao Tong University, Shanghai, China\\
    \textsuperscript{\rm 5}Department of Computer Science and Software Engineering, The University of Western Australia, Australia\\
    s218161821@deakin.edu.au, zhouqianyu@jlu.edu.cn, shaod@pcl.ac.cn, yanghao.cs@sjtu.edu.cn, ye.zhu@deakin.edu.au, richard.dazeley@deakin.edu.au, bruce.lu@uwa.edu.au 
}

\usepackage{bibentry}

\begin{document}

\maketitle

\begin{abstract}
Domain adaptive point cloud completion (DA PCC) aims to narrow the geometric and semantic discrepancies between the labeled source and unlabeled target domains. Existing methods either suffer from limited receptive fields or quadratic complexity due to using CNNs or vision Transformers. In this paper, we present the first work that studies the adaptability of State Space Models (SSMs) in DA PCC and find that directly applying SSMs to DA PCC will encounter several challenges: directly serializing 3D point clouds into 1D sequences often disrupts the spatial topology and local geometric features of the target domain. Besides, the overlook of designs in the learning domain-agnostic representations hinders the adaptation performance. To address these issues, we propose a novel framework, DAPointMamba for DA PCC, that exhibits strong adaptability across domains and has the advantages of global receptive fields and efficient linear complexity. It has three novel modules. 
In particular, Cross-Domain Patch-Level Scanning introduces patch-level geometric correspondences, enabling effective local alignment. Cross-Domain Spatial SSM Alignment further strengthens spatial consistency by modulating patch features based on cross-domain similarity, effectively mitigating fine-grained structural discrepancies. Cross-Domain Channel SSM Alignment actively addresses global semantic gaps by interleaving and aligning feature channels. Extensive experiments on both synthetic and real-world benchmarks demonstrate that our DAPointMamba outperforms state-of-the-art methods with less computational complexity and inference latency.
\end{abstract}


\section{Introduction}
3D point cloud completion (PCC) is a fundamental task in 3D vision, essential for various real-world applications such as autonomous driving, robotics, and virtual reality. Its objective is to recover complete shapes from partial input point clouds accurately. While recent supervised methods ~\cite{tesema2023point,fei2022comprehensive,yu2021pointr, wang2024pointattn} have achieved great progress, their performance often drops significantly when directly applied to unseen datasets due to substantial distribution shifts induced by different sensors, shapes, and \emph{etc,} across domains.

\begin{figure*}[t]
    \centering
    \vspace{-3mm}
    \includegraphics[width=1.0\linewidth]{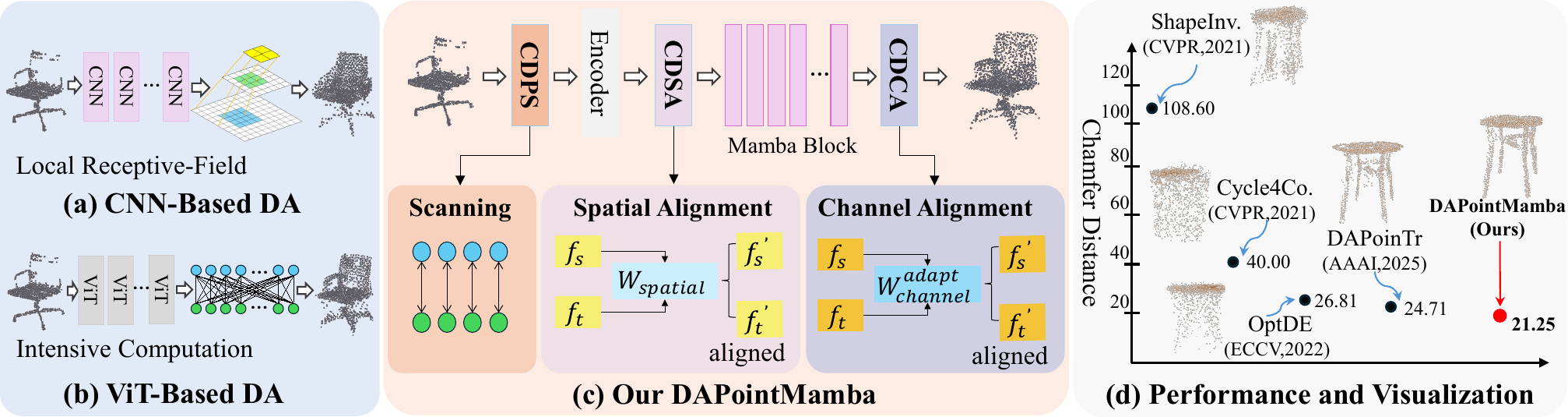}
    \caption{\textbf{Left}: 
    Previous domain adaptive point cloud completion works suffer from limited receptive fields (a) or quadratic complexity (b) due to using CNNs or vision Transformers. \textbf{Middle:} In contrast, we propose a novel framework, DAPointMamba (c), that consists of Cross-Domain Patch-Level Scanning (CDPS), Cross-Domain Spatial SSM Alignment (CDSA), and Cross-Domain Channel SSM Alignment (CDCA). Our model exhibits strong adaptability across domains and has the advantages of global receptive fields and efficient linear complexity. \textbf{Right:} Visualization of the Table category from widely-used 3D-FUTURE benchmark, and our DAPointMamba demonstrates superior domain adaptability (lower Chamfer Distance is better) against state-of-the-art methods. }
    \label{Architecture_Comparison}
    \vspace{-4mm}
\end{figure*}

To address this issue, unsupervised domain adaptation (UDA) techniques are introduced to PCC to narrow the domain shifts between the labeled source domain and the unlabeled target domain. The existing UDA PCC methods can be mainly categorized into several types:
adversarial learning~\cite{chen2019unpaired,zhang2021unsupervised,yang2024syn},data reconstruction~\cite{wen2021cycle4completion,liu2024cloudmix}, disentangled learning~\cite{gong2022optimization}, and self-supervised learning~\cite{hong2023acl,yang2024syn}. Nevertheless, most of them are built upon convolution-based backbones (Figure~\ref{Architecture_Comparison}(a)) and suffer from limited receptive fields and struggle to model global geometric structures during the learning of domain-invariant features for adaptation. Recently, DAPoinTr~\cite{li2025dapointr} introduces the Transformer architecture~\cite{yu2021pointr} in UDA PCC to align the sequence-wise features at the global level and the local level. However, the quadratic complexity of the attention mechanism (Figure~\ref{Architecture_Comparison}(b)) drastically leads to unacceptable computational inefficiency and memory overhead, especially for long sequence patches.

Recently, Mamba, one of emerging State Space Models (SSMs), is capable of modeling long-range dependencies with a global receptive field~\cite{gu2023mamba,liu2024vmamba}. Moreover, Mamba stands out due to its remarkable efficiency, achieving linear computational complexity compared to Transformer-based methods. Recent studies ~\cite{liang2024pointmamba,han2024mamba3d} have validated Mamba's efficacy in point cloud tasks, demonstrating strong capabilities for efficient sequence modeling. Despite its gratifying progress, it is non-trivial to directly apply existing Mamba-based approaches to UDA PCC due to the lack of consideration of domain shifts and tailored designs.
Therefore, how to improve the transferability of Mamba-based models in UDA PCC still remains a very critical and open problem.

In this paper, our goal is to enhance the transferability of Mamba-like models across domains in UDA point cloud completion. Our motivations mainly lie in two aspects. 
Firstly, directly serializing sparse, unstructured 3D point clouds into 1D sequences often disrupts the spatial topology and local geometric features of point clouds, undermining the completion performance in the target domain. For instance, a rule-based scanning strategy may miss critical geometric local details under specific viewpoints, and spatial relationships between points may not be preserved after serialization.
Secondly, current Point Mamba architectures lack tailored designs for learning domain-invariant features, making them sensitive to domain shifts and consequently resulting in poor completion results in target domains.


In this paper, we propose DAPointMamba, the first Mamba-based UDA framework (Figure~\ref{Architecture_Comparison}(c)) explicitly tailored for domain adaptive point cloud completion. Our  DAPointMamba has the advantages of strong transferability toward the target domain, linear computational complexity, and a global receptive field in completion point clouds. Our DAPointMamba comprises three key components. Firstly, Cross-Domain Patch-Level Scanning (CDPS) is introduced to explicitly ensure spatial correspondence across domains by partitioning incomplete point clouds into aligned local patches using shared coordinate normalization and Z-order serialization. Secondly, Cross-Domain Spatial SSM Alignment (CDSA) is presented to explicitly address fine-grained spatial misalignment at the patch level through similarity-guided modulation, reinforcing local feature alignment and effectively mitigating structural discrepancies. Finally, Cross-Domain Channel SSM Alignment (CDCA) is designed to tackle global semantic misalignment by adaptively mixing and modulating feature channels from both source and target domains. Extensive experiments are conducted on diverse benchmarks, including real-world scans (KITTI, ScanNet, and MatterPort3D) and synthetic datasets (3D-FUTURE and ModelNet), showing the effectiveness of DAPointMamba. 

In summary, our contributions are three-fold:

\begin{itemize}
\item We propose DAPointMamba, a novel Mamba-based framework, for UDA point cloud completion. Our framework exhibits strong adaptability across domains and advantages of global receptive fields and linear complexity.

\item We design Cross-Domain Patch-Level Scanning to ensure patch-level geometric correspondences, and Cross-Domain Spatial SSM Alignment and Cross-Domain Channel SSM Alignment to enforce fine-grained spatial alignment and global semantic alignment.



\item Extensive experiments with visualization on both synthetic and real-world datasets consistently confirm the superiority of our approach to state-of-the-art methods in terms of performance and computational efficiency.
\end{itemize}

\begin{figure*}[t]
    \centering
    \includegraphics[width=1.0\linewidth]{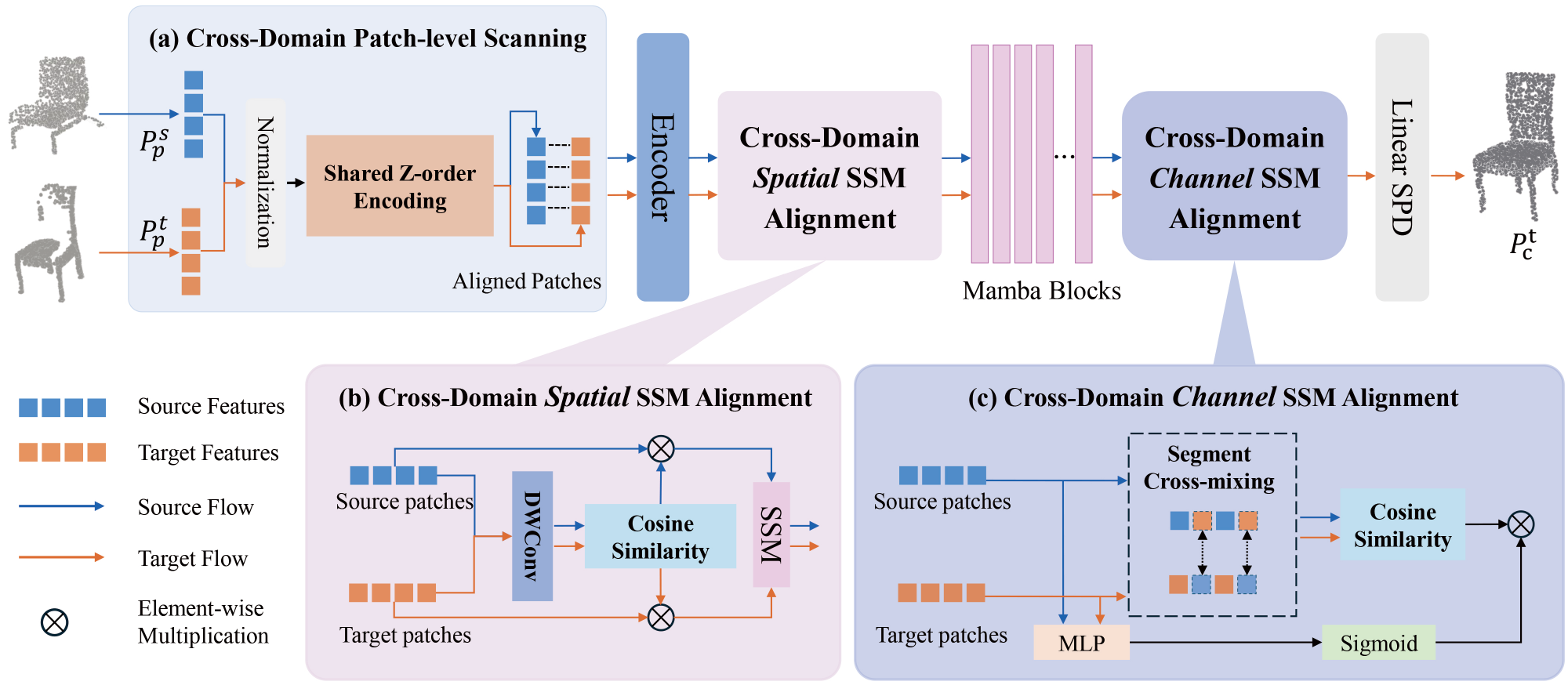}
    \caption{The framework of DAPointMamba for cross-domain point cloud completion, including three key components: (a) Cross-Domain Patch-Level Scanning is designed to close domain shifts by creating spatial correspondence at the patch level. (b) Cross-Domain Spatial SSM Alignment is proposed to solve fine-grained spatial discrepancies across domains. (c) Cross-Domain Channel SSM Alignment is presented to address the semantic structure of feature channels through cross-domain channel mixing and similarity-based modulation. } 
    \label{framework}
\end{figure*}

\section{Related Work}
\subsection{Domain Adaptive Point Cloud Completion}
Point cloud completion (PCC) aims to generate complete shapes from partial input. Early CNN‑based methods~\cite{Yuan2018PCN,Groueix2018AtlasNet,tesema2023point,fei2022comprehensive}  used folding decoders and convolutional architectures to generate coarse shapes. Later advancements~\cite{Tchapmi2019TopNet,xiang2021snowflakenet} refined the output resolution and structural detail with hierarchical or progressive generation strategies. 
Transformer-based models~\cite{yu2021pointr, zhou2022seedformer} improve global feature modeling but suffer from a high computational cost. Unfortunately, PCC models often degrade when transferred across different domains. UDA has thus been explored to bridge this gap via
adversarial learning~\cite{chen2019unpaired,zhang2021unsupervised}, data reconstruction~\cite{wen2021cycle4completion,liu2024cloudmix}, disentangled learning~\cite{gong2022optimization}, and self-supervised learning~\cite{hong2023acl,yang2024syn}. Nonetheless, existing methods either suffer from limited receptive fields or quadratic complexity due to using CNNs or Transformers. Thus, it is an urgent need to investigate UDA PCC with global modeling, linear complexity, and high transferability toward the target domain.

\subsection{State Space Models (SSMs)}
Mamba, as well as State space models (SSMs)~\cite{gu2023mamba}, have recently garnered increasing attention due to their strong capabilities in modeling long-range dependencies with linear computational complexity. VMamba and Vision Mamba~\cite{liu2024vmamba,zhu2024vision} introduce Mamba architecture in vision tasks. 
Recently, a series of works has been inspired to study Mamba in point cloud analysis, where the representative works are PointMamba, Point Cloud Mamba, Mamba3D, \emph{etc}~\cite{liang2024pointmamba,zhang2025point,han2024mamba3d,bahri2025spectral,wang2024pointramba,Li2025LGMamba}. In the completion task, 3DMambaComplete~\cite{li20243dmambacomplete} incorporates Mamba blocks to enhance local details and global structure, while it also incorporates cross-attention mechanisms and is not a pure Mamba design.
To the best of our knowledge,  this is the first work that studies Mamba's adaptability in UDA PCC. 



\section{Methodology}
In this section, we present DAPointMamba, a novel Mamba-based architecture specifically designed for domain-adaptive point cloud completion. DAPointMamba demonstrates superior adaptability to unseen domains while maintaining the advantages of global receptive fields and efficient linear complexity. 
As illustrated in Figure~\ref{framework}, our framework comprises three key components: Cross-Domain Patch-Level Scanning (CDPS), Cross-Domain Spatial SSM Alignment (CDSA), and Cross-Domain Channel SSM Alignment (CDCA). 
Specifically, CDPS is designed to spatially align point patches, mitigating domain discrepancies between source and target domains during the serialization process.
Furthermore, CDSA explicitly models and strengthens spatial correspondences across domain-specific features at the patch level, effectively aligning spatial information. 
Complementarily, CDCA addresses global semantic misalignment by adaptively mixing and modulating feature channels across domains, enabling robust and consistent alignment of both spatial and semantic representations between source and target domains.

\subsection{Cross-Domain Patch-Level Scanning}
We have noticed that cross-domain point cloud completion faces severe geometric inconsistencies between source and target domains, where direct serialization of sparse, unstructured 3D point clouds into 1D sequences often disrupts spatial topology and undermines local geometric features. Recent works~\cite{wei2022learning,wei2024multi} have demonstrated that local geometric structures and patch-level context are crucial for both capturing fine-grained details and achieving robust adaptation. However, most existing methods~\cite{liang2024pointmamba, wu2024point} rely on global or static grouping strategies, which ignore domain-specific variations and fail to guarantee spatial correspondence across domains during serialization.

To address this, we propose a cross-domain patch-level scanning strategy with shared Z-order curve serialization named \emph{Cross-Domain Patch-Level Scanning} (CDPS). Unlike heuristic global scanning~\cite{liang2024pointmamba,wu2024point}, CDPS ensures that local patches in the source domain and target domain are spatially aligned, thereby facilitating more effective feature adaptation across domains. Specifically, given a sample \( X_s, X_t \in \mathbb{R}^{B \times N \times 3} \),  from source and target domains, respectively, we first normalize and compute the shared minimum coordinate across both domains:
\begin{equation}
    C_{min}\!=\!min(min(X_s,\!dim=1), min(X_t,\!dim=1))
\end{equation}
Each normalized point cloud is then discretized into a shared grid space and serialized into a 1D sequence using a consistent Z-order curve encoding, which maps 3D coordinates into a 1D sequence while maintaining spatial proximity:
\begin{equation}
     G_s = [X_s-C_{min} * scale], \quad
     G_t = [X_t-C_{min} * scale] 
\end{equation}
\begin{equation}
    Z_s = {\text{Z-order}}(G_s), Z_t = {\text{Z-order}}(G_t)
\end{equation}
We sort the points from both domains according to their Z-order values and partition the sequence into $G$ patches, each containing $K$ points: 
\begin{equation}
\begin{aligned}
    X_s = \{P_s^{(1)},P_s^{(2)},..., P_s^{(G)}\}, \\
    X_t = \{P_t^{(1)},P_t^{(2)},..., P_t^{(G)}\}
\end{aligned}
\end{equation}
where \(P_s^{(g)},P_t^{(g)} \in \mathbb{R}^{K \times 3}\) denote the $g$-th patch in the source and target domains, respectively. Due to the unified normalization and Z-order serialization, the $g$-th patch in both domains corresponds to the same region, enabling precise patch-level alignment.

This patch-level alignment guarantees that each pair of patches consistently covers an identical spatial region across domains, thus providing a reliable basis for patch-level feature alignment and cross-domain adaptation. CDPS explicitly preserves spatial consistency during serialization, fundamentally mitigating the mismatch issues inherent in traditional global or purely sequential approaches.

\subsection{Cross-Domain Spatial SSM Alignment}
While the CDPS module effectively aligns feature distributions at the patch level, residual fine-grained discrepancies within corresponding local patches persist, limiting the precise alignment of local geometric details and ultimately hindering domain adaptability to unseen domains.

To solve this problem, we introduce the \emph{Cross-Domain Spatial SSM Alignment} (CDSA) module. The CDSA effectively reinforces the alignment of spatially corresponding local patches while preserving structural diversity in regions with significant geometric variation. Specifically, given patch-level features $X_s, X_t \in R^{B\times D\times G}$, where $B$ is the batch size, $D$ is the feature dimension, and $G$ is the number of patches, the Spatial SSM module operates as follows:
\begin{equation}
    \mathcal D_s = DWConv(P_s^{G}), \quad
    \mathcal D_t = DWConv(P_t^{G}) 
\end{equation}
where $DWConv(\cdot)$ denotes a depthwise 1D convolution applied along the patch dimension. 
For each patch $g$ in the batch, we compute the cosine similarity between the convolved features to obtain the space similarity weight $\mathcal{W_{\text{spatial}}}$, which can be represented as $\mathcal W_{spatial} = cos(D_s, D_t)$. 
Finally, we modulate the patch features accordingly:
\begin{equation}
\begin{aligned}
    \tilde{X}_s = P_s^G \odot W_{spatial}, \quad
    \tilde{X}_t = P_t^G \odot W_{spatial}    
\end{aligned}
\end{equation}
where $\tilde{X}_s$ and $\tilde{X}_t$ denote the patch features of the source and target domains, respectively.  
To explicitly encourage local feature consistency across domains, we further compute a mean squared error (MSE) loss between $\tilde{X}_s$ and $\tilde{X}_t$:
\begin{equation}
    \mathcal L_{sp} = \frac{1}{BDG} \sum_{b=1}^{B} \sum_{d=1}^{D} \sum_{g=1}^{G} \Big( \tilde{X}_{s,b,d,g} - \tilde{X}_{t,b,d,g} \Big)^2
\end{equation}
This design significantly improves the model's capacity to capture fine-grained local details by effectively modulating patch features based on cross-domain spatial similarities, enhancing its ability to generalize to unseen domains.

\subsection{Cross-Domain Channel SSM Alignment}
Despite effective local alignments from CDSA, global semantic inconsistencies across domains remain a critical challenge due to significant domain gaps. Meanwhile, Mamba-based architectures have demonstrated impressive efficiency and scalability for long-sequence modeling with linear computational complexity. However, recent studies~\cite{liang2024pointmamba,han2024mamba3d} indicate that Mamba’s inherently one-dimensional and causal formulation limits its capability to capture complex global interactions across feature channels, particularly when dealing with unordered and high-dimensional point cloud data. Consequently, Mamba-based backbones often exhibit insufficient global feature interaction and suboptimal performance in addressing cross-domain semantic discrepancies, especially under severe distribution shifts between source and target domains.

To mitigate these limitations, we propose the \emph{Cross-Domain Channel SSM Alignment} (CDCA), which explicitly enhances global feature consistency at the channel level across domains. As illustrated in Figure~\ref{framework} (c), we first compute the global feature for each domain:
\begin{equation}
    g_s = \frac{1}{G} \sum_{g=1}^{G} X_s \in \mathbb{R}^{B \times D},  \\
    g_t = \frac{1}{G} \sum_{g=1}^{G} X_t \in \mathbb{R}^{B \times D}
\end{equation}
Then, the global features are concatenated and passed through a learnable alignment strength estimator:
\begin{equation}
     \alpha = \mathrm{Sigmoid}\left( \mathrm{MLP}\left( [g_s, g_t] \right) \right) \in \mathbb{R}^{B \times 1}
\end{equation}
The feature channels are first divided into $S$ non-overlapping segments for both the source and target domain representations, $X_s$ and $X_t$, respectively.
To reduce the domain gap and encourage information exchange, we construct mixed channel representations $X_{s,mix}$ and $X_{t,mix}$ by interleaving segments from both domains:
\begin{equation}
\begin{aligned}
    X_{s,mix} = [X_s^{(1)}, X_t^{(2)}, X_s^{(3)}, X_t^{(4)},\cdots], \\
    X_{t,mix} = [X_t^{(1)}, X_s^{(2)}, X_t^{(3)}, X_s^{(4)},\cdots]
\end{aligned}  
\end{equation}
We then compute the cosine similarity between the mixed representations to derive a channel-wise alignment weight, which can be represented as:
\begin{align}
    \mathcal W_{channel} = cos(X_{s,mix},X_{t,mix}) \in \mathbb{R}^{B \times G}
\end{align}
And the similarity weights $W_{channel}$ are modulated by the estimated alignment strength $\alpha$ to generate adaptive similarity  $\tilde{W}_{channel}$. Finally, the original features are modulated by the adaptive similarity weights $\tilde{W}_{channel}$:
\begin{equation}
     \tilde{F}_s = X_s \odot \tilde{W}_{channel},
    \tilde{F}_t = X_t \odot \tilde{W}_{channel}   
\end{equation}
To explicitly regularize global feature alignment, we minimize the mean squared error between the modulated $\tilde{F}_s$ and $\tilde{F}_t$ from source and target features, respectively:
\begin{equation}
    \mathcal L_{ch} = \frac{1}{BDG} \sum_{b=1}^{B} \sum_{d=1}^{D} \sum_{g=1}^{G} \Big( \tilde{F}_{s,b,d,g} - \tilde{F}_{t,b,d,g} \Big)^2
\end{equation}

By explicitly modeling and aligning global channel-wise information, the Channel SSM compensates for the inherent limitations of the Mamba backbone in global context aggregation. This design ensures that high-level semantic structures remain consistent across domains, which is crucial for robust domain adaptation point cloud completion. Therefore, the total loss function can be formulated as:
\begin{equation}
    \mathcal L_{total} = Loss_{(CD)} + \lambda L_{sp} + \beta L_{ch}
\end{equation}

\section{Experiments}
\subsection{Experimental Settings}
\label{Section4.1}
\noindent\textbf{Datasets.} Following previous standard protocol in UDA PCC (Chen, Chen, and Mitra 2019; Li et al. 2025b), we evaluate the effectiveness of our proposed DAPointMamba and compare it with state-of-the-art models on three commonly used datasets. We use 3D data from CRN~\cite{wang2020cascaded} as our source domain, and the datasets, including Real-World Scans, 3D-FUTURE~\cite{fu20213d}, and ModelNet~\cite{wu20153d} as our target domains. For the CRN dataset, we use 26,863 samples drawn from categories shared between CRN and the other datasets for DA. For the real-world scan datasets, we evaluate our model on ScanNet~\cite{dai2017scannet}, MatterPort3D~\cite{chang2017matterport3d}, and KITTI~\cite{geiger2012we}. We select shared categories between source and target domains, and all input scans are uniformly downsampled to 2048 points for unpaired training and inference.

\noindent\textbf{Implementation Details.}
All experiments are conducted on an RTX 4090 with 64GB RAM. We use refinement modules of PointMamba~\cite{liang2024pointmamba} as the backbone of our UDA PCC network. For the training, we employ an initial learning rate of $1 \times 10^{-3}$ and a weight decay of $5 \times 10^{-2}$. The batch size is set to 32. To balance losses, weights of $\lambda$ and $\beta$ are set to 0.1 and 0.1, respectively. Following prior works, we adopt the Unidirectional Chamfer Distance (UCD), Unidirectional Hausdorff Distance (UHD), and Chamfer Distance (CD) as evaluation metrics. 
\begin{table}[t!]
\centering
\footnotesize
\setlength{\tabcolsep}{1mm}
\begin{tabular}{l|l|cccccc}
\toprule
Methods&Avg&Cabinet&Chair&Lamp&Sofa&Table \\
\midrule
Pcl2Pcl&92.83&57.23&43.91&157.86&63.23&141.92\\
ShapeInv.&53.21 &38.54&26.30&48.57&44.02&108.60\\
Cycle4Comp&45.39 &32.62&34.08&77.19&43.05&40.00\\
ACL-SPC&35.97&70.12&23.87&31.75&28.74&25.38\\
OptDE&28.99&28.37&21.87&29.92&37.98&26.81 \\
DAPoinTr&22.35 &\textbf{18.46} &17.60 &27.91 &23.08 &24.71\\
DAPointMamba&\textbf{20.40}&19.35&\textbf{16.21}&\textbf{22.81}&\textbf{22.38}&\textbf{21.25}\\
\bottomrule
\end{tabular}
\caption{Cross-domain completion results on 3D-FUTURE.  Chamfer Distance (CD)$\downarrow$ is used as the metric, and the scale factor is $10^4$. Lower is better. }
\label{table1}
\end{table}

\subsection{Main Results}
\label{4.2}
In this section, we conduct comprehensive experimental comparisons on widely used benchmarks spanning both real-world and synthetic domains. Specifically, we evaluate our method on real-world scans from KITTI, ScanNet, and MatterPort3D, as well as synthetic datasets including 3D-FUTURE and ModelNet, to thoroughly assess its performance across diverse scenarios.

\noindent \textbf{Comparison Methods.} We compare our method with  state-of-the-art approaches in UDA PCC. These include Pcl2Pcl~\cite{chen2019unpaired}, ShapeInversion~\cite{zhang2021unsupervised}, Cycle4Completion~\cite{zhang2021unsupervised}, OptDE~\cite{gong2022optimization}, ACL-SPC~\cite{hong2023acl} and DAPoinTr~\cite{li2025dapointr}.

\noindent \textbf{Results on 3D-FUTURE benchmark.} As shown in Table \ref{table1}, our proposed method DAPointMamba achieves consistent and significant improvements over the state-of-the-art UDA PCC methods on the 3D-FUTURE benchmark. Compared with the second-best method, DAPoinTr, our approach achieves an average improvement of 1.95 (from 22.35 to \textbf{20.40}) in Chamfer Distance (CD) across all categories. Notably, our model brings substantial improvements in challenging categories such as lamp and table, with remarkable margins (see the 5th and 7th columns in Table \ref{table1}). Although the result on the cabinet is slightly lower than DAPoinTr, the overall performance proves that our model can better align feature distributions between source and target domains, while effectively leveraging the capacity of Mamba
to capture both local and global geometric structures.

\begin{table}[t!]
\centering
\footnotesize
\small
\setlength{\tabcolsep}{0.5mm}
\begin{tabular}{l|l|cccccc}
\toprule
Methods&Avg&Plane&Car&Chair&Lamp&Sofa&Table \\
\midrule
Pcl2Pcl&68.14&18.53&17.54&43.58&126.80&38.78&163.62\\
ShapeInv.& 41.61 &3.78& 15.66& 22.25& 60.42&22.25 & 125.31\\
Cycle4Comp& 28.65 &5.77& 11.85& 26.67& 83.34& 22.82& 21.47 \\
ACL-SPC&34.89&5.75&11.73&43.08&106.29&25.62&16.89\\
OptDE& 15.94 &\textbf{2.18}& 9.80& 14.71& 39.74& 19.43& \textbf{9.75} \\
DAPoinTr&13.79 &2.38 &8.04 &13.83 &33.26 &12.72 &12.51\\
DAPointMamba&\textbf{13.11}&2.30&\textbf{7.58}&\textbf{13.15}&\textbf{32.04}&\textbf{12.48}&11.08\\
\bottomrule
\end{tabular}
\caption{Cross-domain completion results on ModelNet.  Chamfer Distance (CD)$\downarrow$ is used as the metric, and the scale factor is $10^4$. Lower is better.}
\label{table2}
\end{table}

\begin{table}[t]
\centering
\footnotesize
\setlength{\tabcolsep}{0.5mm}
\begin{tabular}{@{}l|cc|cc|cc@{}}
\toprule
\multirow{2}{*}{Methods}
&\multicolumn{2}{c}{ScanNet}&\multicolumn{2}{c}{MatterPort3D}&\multicolumn{2}{c}{KITTI}\\
\cmidrule(lr){2-7}
&Chair&Table&Chair&Table&Car \\
\midrule
Pcl2Pcl&17.3/10.1&9.1/11.8&15.9/10.5&6.0/11.8&9.2/14.1\\
ShapeInv.&3.2/10.1 &3.3/11.9&3.6/10.0&3.1/11.8&2.9/13.8\\
Cycle4Comp&5.1/6.4 &3.6/5.9&8.0/8.4&4.2/6.8&3.3/5.8\\
ACL-SPC&1.4/4.7& 1.8/5.1&1.8/4.8&2.1/4.9&2.0/4.9&\\
OptDE&2.6/5.5&1.9/4.6&3.0/5.5&1.9/5.3&1.6/3.5 \\
DAPoinTr&1.1/\textbf{2.7} &0.96/2.7 &1.3/\textbf{2.9} &1.2/\textbf{2.8}&0.45/\textbf{1.8} \\
DAPointMamba&\textbf{0.95}/2.8&\textbf{0.91}/\textbf{2.6}&\textbf{1.1}/3.0&\textbf{1.0}/3.0&\textbf{0.40}/2.1&\\
\bottomrule
\end{tabular}
\caption{Cross-domain completion results on real-world scans. [UCD$\downarrow$/UHD$\downarrow$] are used as the metrics, and the scale factor is $10^4$ and $10^2$ for UCD (Unidirectional Chamfer Distance) and UHD (Unidirectional Hausdorff Distance), respectively.  A lower value of UCD or UHD is better. }
\label{table3}
\end{table}

\begin{table*}[t]
\centering
\footnotesize
\begin{tabular}{l|ccc|ccccc|c}
\toprule
Methods&CDPS&CDSA&CDCA&Cabinet&Chair&Lamp&Sofa&Table&Avg \\
\midrule
Baseline&&&&23.14&17.21&26.68&25.71&24.16&23.38\\
\midrule
\multirow{3}{*}{ DAPointMamba}
&\checkmark&&&20.06&17.02&25.37&23.35&22.85&21.73 \\
&\checkmark&\checkmark&&19.68&16.45&24.29&22.80&22.61&21.17 \\
&\checkmark&\checkmark&\checkmark&\textbf{19.35}&\textbf{16.21}&\textbf{22.81}&\textbf{22.38}&\textbf{21.25}&\textbf{20.40} \\
\bottomrule
\end{tabular}
\caption{Ablation studies on each proposed module of our DAPointMamba on 3D-FUTURE dataset. Chamfer Distance (CD)$\downarrow$ is used as the metric, and the scale factor is $10^4$. Lower is better. }
\label{table4}
\end{table*}
\noindent \textbf{Results on ModelNet benchmark.}  As presented in Table \ref{table2}, our DAPointMamba demonstrates superior average performance compared to the state-of-the-art methods. Specifically, DAPointMamba achieves the best performance across all categories compared with the current best model, DAPoinTr. This superiority can be attributed to our innovative cross-domain alignment designs, which explicitly address local geometric inconsistencies and global semantic variations. Our innovative designs can effectively mitigate domain shifts and narrow domain gaps at the patch and global level. Such consistent improvements highlight the robustness and domain adaptability of our proposed architecture when tackling complex UDA PCC tasks.

\noindent \textbf{Results on Real-World Scans benchmark.} Table~\ref{table3} presents quantitative results on three widely used real-world scan datasets: ScanNet, MatterPort3D, and KITTI. We evaluate performance using both Unidirectional Chamfer Distance (UCD) and Unidirectional Hausdorff Distance (UHD), which respectively measure overall geometric fidelity and worst-case point-wise deviation. As shown in Table~\ref{table3}, DAPointMamba consistently outperforms all prior methods across all datasets in terms of the UCD metric, indicating our approach can preserve global shape structures and align feature distributions across domains. While our UHD performance is slightly inferior to previous methods, we consider that this is due to the nature of the UHD metric, which emphasizes the maximum point-wise error and is highly sensitive to isolated boundary outliers. Since DAPointMamba is designed to optimize holistic shape prediction through consistent local-global alignment, rather than minimizing extreme outliers, it is naturally more aligned with the UCD objective. 

\subsection{Ablation Studies}
In this section, we take a closer look at the effects of the designed components of our model, including the Cross-Domain Patch-Level Scanning (CDPS), Cross-Domain Spatial SSM Alignment (CDSA), and Cross-Domain Channel SSM Alignment (CDCA). As shown in Table 4, we conduct all ablation experiments on the 3D-FUTURE dataset, using the CRN dataset as the source domain and 3D-FUTURE as the target domain.

\noindent \textbf{Effectiveness of each component.} We begin with a \textbf{Baseline} that independently serializes source and target point clouds using the Z-order curve without considering cross-domain spatial correspondence. As shown in Table~\ref{table4}, this naive strategy yields suboptimal performance, with an average Chamfer Distance (CD) of 23.38, due to the misalignment of spatial structures and the inability to learn domain-invariant features. Introducing the \textbf{CDPS} leads to a substantial improvement, reducing the average CD to 21.73. This module establishes spatial correspondence by aligning patches within a shared coordinate space, effectively mitigating local geometric discrepancies. Notably, the cabinet and sofa categories exhibit marked gains, with CD values dropping from 23.14 to 20.06 and from 25.71 to 23.35, respectively. Building on this, we incorporate the \textbf{CDSA}, which explicitly models patch-level similarities between source and target domains. This further reduces the average CD to 21.17. The improvement is particularly evident in geometrically complex categories such as lamp (from 25.37 to 24.29) and chair (from 17.02 to 16.45), highlighting CDSA’s ability to handle fine-grained local misalignments. Finally, by adding \textbf{CDCA}, we achieve an average CD of 20.40—the best overall performance. CDCA addresses residual domain shifts in the global semantic space by adaptively aligning feature channels. Its effectiveness is especially pronounced in high-variance categories such as lamp and table, where CD drops from 24.29 to 22.81 and from 22.61 to 21.25, respectively. These results demonstrate that all individual components are complementary, and their integration leads to significant improvements in UDA point cloud completion.

\begin{table}[t]
    \centering
    
    \begin{tabular}{c|ccc}
    \toprule
         Model& Params(M)&FLOPs(G)&Time(ms) \\
         \midrule
         DAPoinTr& 36.904 & 24.912&23.774 \\
         \midrule
         DAPointMamba&\textbf{9.571}&\textbf{5.192}&\textbf{3.820}\\
         \bottomrule
    \end{tabular}
    \caption{Computational efficiency of the current state-of-the-art approach DAPoinTr and our DAPointMamba. }
    \label{table5}
\end{table}

\begin{figure*}[t]
\centering
\includegraphics[width=1.0\textwidth]{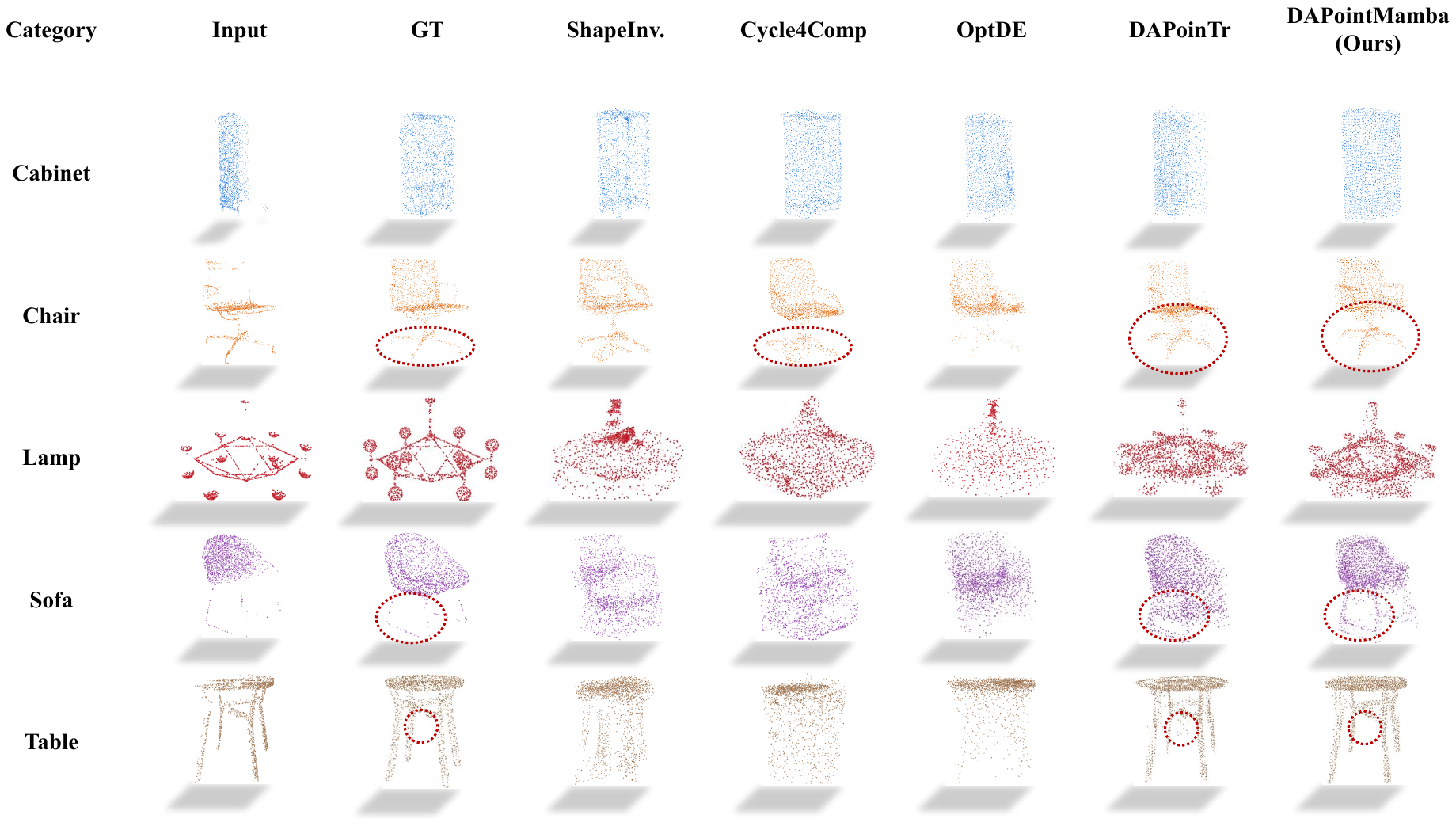} 
\caption{Visualization comparisons between ours and state-of-the-art PCC methods on 3D-FUTURE dataset. }
\label{3D-FUTURE-Visualization}
\end{figure*}

\begin{figure}
    \centering
    \includegraphics[width=1\linewidth]{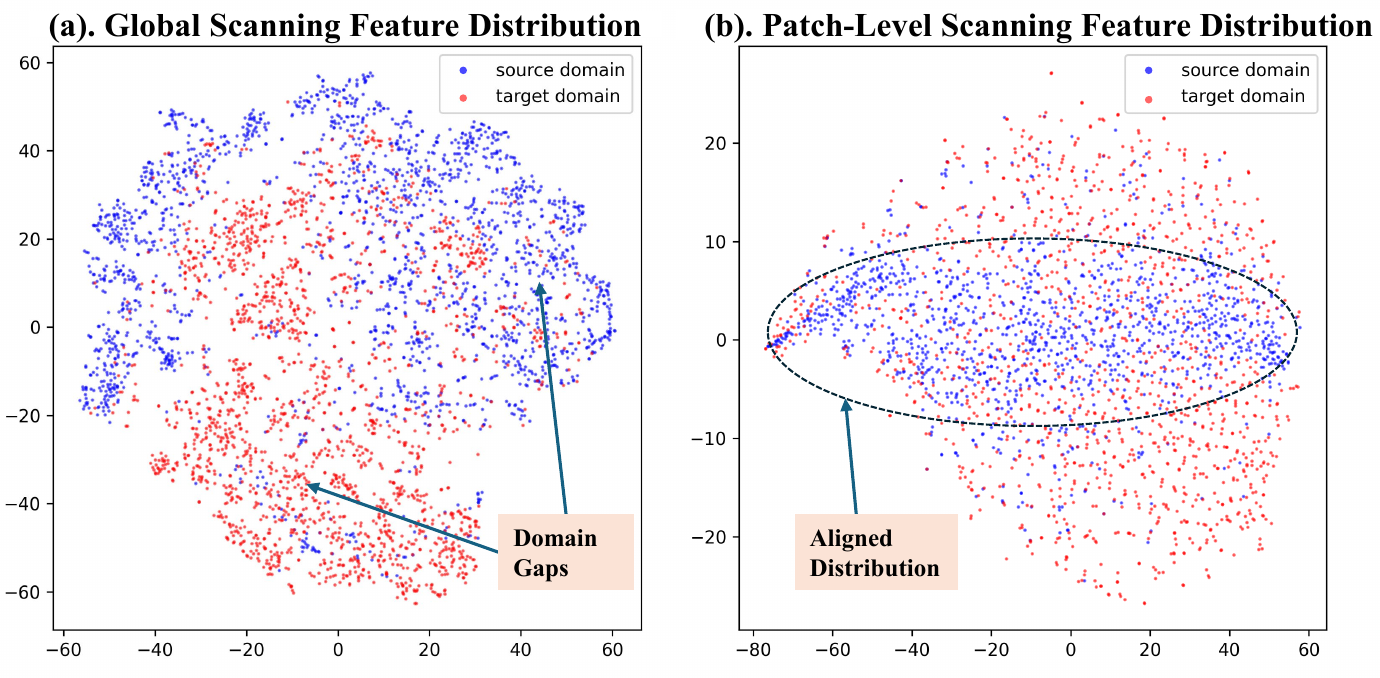}
    \captionof{figure} {Visualization of the feature distribution of Cross-Domain Patch-Level Scanning (CDPS) compared with the global grouping scan strategy.}
    \label{CDPS_Feature_Distribution}
\end{figure}

\subsection{Visualization and Analysis}
\noindent \textbf{Computational Efficiency.} To assess the computational efficiency of our proposed DAPointMamba, we conduct experiments on 3D-FUTURE and compare it with the state-of-the-art DAPoinTr from different perspectives, including model parameters, floating-point operations per second(FLOPs), and inference time. For fair comparison, all evaluations are conducted on an RTX 4090 GPU with 64GB RAM. We fix the batch size to 1 and the number of points to 2048, and the inference time is averaged over 100 experiments to ensure stability. As shown in Table \ref{table5}, our DAPointMamba exhibits significantly lower computational cost compared to the current best model, DAPoinTr. Our model greatly reduces parameters (9.571M vs. 36.904M), FLOPs (5.192G vs. 24.912G), and inference time (3.820 ms vs. 23.774 ms). These results demonstrate that DAPointMamba achieves superior performance with over \textbf{80\%} lower computational cost and latency, making it highly efficient for real-time deployment. 

\noindent \textbf{Visualization results on 3D-FUTURE.} 
Figure~\ref{3D-FUTURE-Visualization} presents visualizations of point cloud completion results on the 3D-FUTURE dataset. Compared to prior state-of-the-art methods, our DAPointMamba consistently predicts more accurate and structurally complete outputs, with a clear advantage in preserving fine-grained local details and global shape. This improvement is particularly evident in the cases of chairs, sofas, and tables, as highlighted in the third, fourth, and last rows of the figure. These visual results further demonstrate our robustness in delivering better reconstructions across various cross-domain scenarios. 

\noindent \textbf{TSNE Visualization of Global and Patch-Level Scanning.} The TSNE visualization (Figure~\ref{CDPS_Feature_Distribution}) compares the effectiveness of our proposed Cross-Domain Patch-Level Scanning (CDPS) strategy with the conventional global scanning approach in terms of feature distribution alignment between source and target domains. As illustrated in Figure~\ref{CDPS_Feature_Distribution}(a), the global scanning method leads to a distinct separation between source (blue points) and target domain (red points), indicating apparent domain gaps and inadequate feature alignment. In contrast, CDPS, as demonstrated in Figure~\ref{CDPS_Feature_Distribution}(b), acquires superior representations, exhibiting significantly improved overlap between source and target domain features. Specifically, the alignment achieved through our CDPS module ensures that features from different domains occupy more consistent and coherent regions in the embedding space. This closer alignment underscores CDPS’s effectiveness in explicitly maintaining spatial correspondences and mitigating local geometric discrepancies at the patch level, thereby facilitating more robust and generalizable cross-domain adaptation.

\section{Conclusion}
In this paper, we present DAPointMamba, a novel and pioneering framework aimed at enhancing the transferability of Mamba blocks for point cloud completion across various domains. Through the introduction of several new modules, including Cross-Domain Patch-level Scanning, Cross-Domain Spatial SSM Alignment, and Cross-Domain Channel SSM Alignment, our DAPointMamba effectively addresses both local and global domain discrepancies, resulting in strong adaptability across domains. Furthermore, compared to the Transformer-based architecture, our DAPointMamba achieves efficient long-sequence modeling with linear complexity. Extensive experiments on synthetic and real-world datasets demonstrate superior performance to state-of-the-art methods, particularly in complex object categories. 
\section{Acknowledgments}
This work was supported by the National Natural Science Foundation of China (Grant No. 62502178).
\bibliography{aaai2026}

\clearpage           


\section*{Supplementary material}

# DAPointMamba: Domain Adaptive Point Mamba for Point Cloud Completion Supplementary Material

The supplementary material presents further details, which are structured as follows:

**Section A** : Visualization and Analysis
    **Section A.1** : Visualization Results of ModelNet
    **Section A.2** : Visualization Results of Real-World Scans
**Section B** : Ablation Study of Different Grid Scales
**Section C** : Limitations and Future Work

## A. Visualization and Analysis

### A.1 Visualization Results of ModelNet.

As illustrated in Figure 1, we visualize the completion results on ModelNet. Compared to the previous state-of-the-art model, DAPoinTr, our method generates more accurate and visually complete point clouds, consistently preserving fine-grained local details. We particularly observed this improvement in the cases of Car, Chair, and Table categories (see the 3rd, 4th, and last row in Figure 1). These results further confirm that DAPointMamba effectively captures both local and global features, achieving strong domain adaptability and robust performance in complex categories.

### A.2 Visualization Results of Real-World Scans

**Visualization Results of KITTI.** In addition, we also visualize the completion results on the real-world dataset KITTI. As depicted in Figure 2, it can be easily observed that our proposed DAPointMamba consistently outperforms previous UDA PCC methods by accurately generating missing parts and maintaining the integrity of complex car structures. This superior performance is attributed to the innovative patch-level alignment and feature channels-based alignment, ensuring that our method better recovers local geometric structures and global shape contexts under challenging real-world scenarios.

**Visualization Results of MatterPort3D.** Figure 3 illustrates qualitative completion results on the MatterPort3D dataset. Compared to existing methods, our model achieves higher-quality completions, exhibiting enhanced preservation and reconstruction of fine-grained local structure. The advantages are particularly pronounced in challenging categories, Chair and Table, where our DAPointMamba generates more precise local geometric details. These visualizations underscore our model's robust domain adaptability and preservation of structural integrity in PCC.

| Scale | Cabinet | Chair | Lamp | Sofa | Table | Avg |
|---|---|---|---|---|---|---|
| 5 | 20.17 | 16.95 | 23.21 | 22.45 | 22.88 | 21.13 |
| 10 | **19.35** | 16.21 | 22.81 | **22.38** | **21.25** | **20.40** |
| 15 | 20.05 | **16.14** | **22.64** | 22.84 | 23.81 | 21.10 |

Table 1: Ablation Study of Different Grid Scales on 3D-FUTURE dataset. We take Chamfer Distance(CD)↓ as the metric, and the scale factor is $10^4$. Lower is better.

## B. Ablation Study of Different Grid Scales

We conduct an ablation study to comprehensively investigate the influence of grid resolution scale within the Cross-Domain Patch-Level Scanning (CDPS) module, a key component for narrow domain gaps at patch level. Specifically, we use CRN and 3D-FUTURE datasets as source and target domains, respectively. As shown in Table 1, we evaluate three different scale settings. The experiment results indicate an explicit dependency of the model's domain adaptation performance on the selected grid resolution. The coarse grid (scale=5) lacks sufficient spatial granularity, resulting in suboptimal performance (CD=21.13). Conversely, adopting a finer scale (scale=15) increases noise sensitivity and achieves only marginal improvement (CD=21.10). Notably, the intermediate scale (scale=10) achieves the optimal balance between spatial granularity and robustness, yielding the best average CD of 20.40. This setting consistently benefits categories with intricate geometries, such as Cabinet, Sofa, and Table. These findings highlight the importance of choosing an appropriate grid resolution to address cross-domain geometric misalignment in PCC effectively.

## C. Limitations and Future Work

Our DAPointMamba achieves state-of-the-art performance in unsupervised domain adaptation for point cloud completion. While our approach can recover high-quality local and global details from partial inputs, it still faces challenges when processing complex structures. Additionally, the fixed grid partitioning strategy may restrict domain adaptability for objects with various scales from different domains. In future work, we would like to explore these.

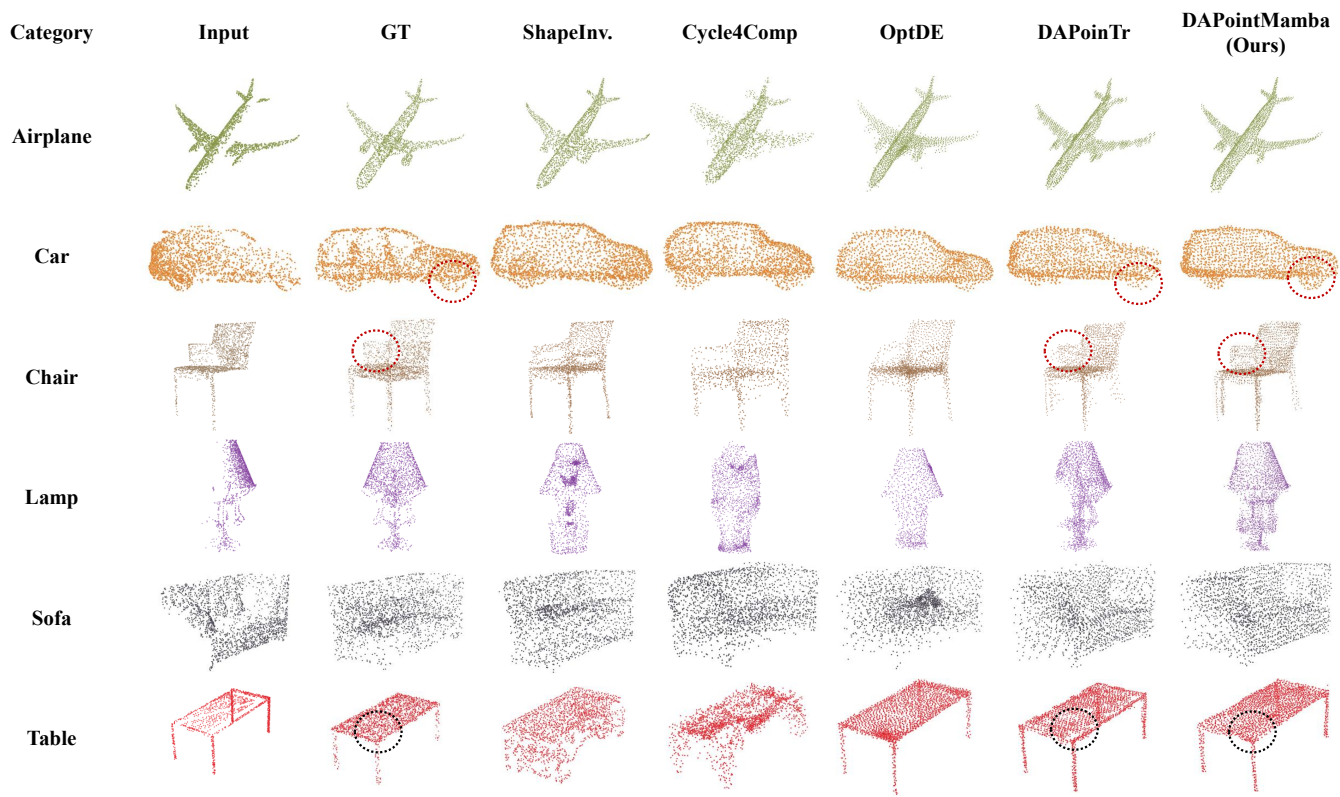

Figure 1: Visualization comparisons with state-of-the-art PCC methods on ModelNet dataset.

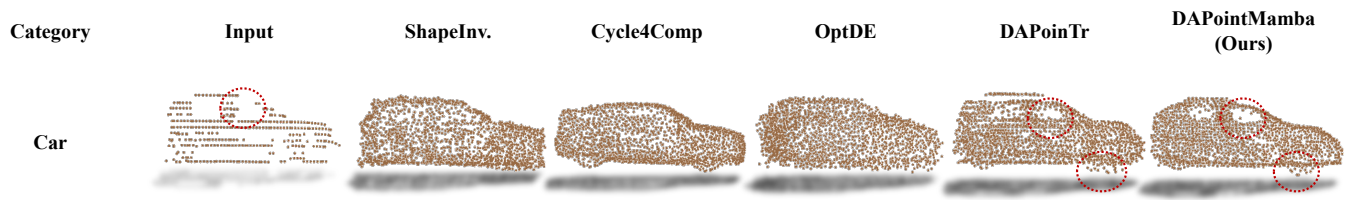

Figure 2: Visualization comparisons with state-of-the-art PCC methods on KITTI dataset.

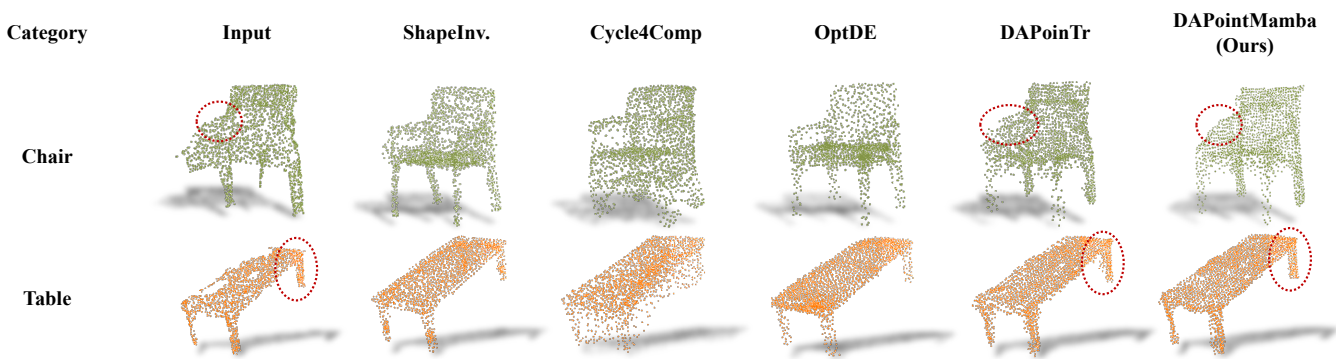

Figure 3: Visualization comparisons with state-of-the-art PCC methods on MatterPort3D dataset.



\end{document}